# Enhanced Integrated Scoring for Cleaning Dirty Texts

**Wilson Wong, Wei Liu and Mohammed Bennamoun**
School of Computer Science and Software Engineering
University of Western Australia
Crawley WA 6009
{wilson, wei, bennamou}@csse.uwa.edu.au

## Abstract

An increasing number of approaches for ontology engineering from text are gearing towards the use of online sources such as company intranet and the World Wide Web. Despite such rise, not much work can be found in aspects of preprocessing and cleaning dirty texts from online sources. This paper presents an enhancement of an **I**ntegrated **S**coring for **S**pelling error correction, **A**bbreviation expansion and **C**ase restoration (ISSAC). ISSAC is implemented as part of a text preprocessing phase in an ontology engineering system. New evaluations performed on the enhanced ISSAC using 700 chat records reveal an improved accuracy of 98% as compared to 96.5% and 71% based on the use of only basic ISSAC and of Aspell, respectively.
**Keywords**: Spelling error correction, abbreviation expansion, case restoration

## 1 Introduction

Ontology is gaining applicability across a wide range of applications such as information retrieval, knowledge acquisition and management, and the Semantic Web. Over the years, manual construction and maintenance of ontologies have become increasingly expensive due to factors such as increases in labour cost, manpower and knowledge fluctuation. These factors have prompted an increasing effort in automatic and semi-automatic engineering of ontologies using information from electronic sources. A particular type of such electronic sources that is becoming popular is texts from the World Wide Web.

The quality of texts from online sources for ontology engineering can vary anywhere between *dirty* and *clean*. On the one hand, the quality of texts in the form of blogs, emails and chat logs can be extremely poor. The sentences in dirty texts are typically full of spelling errors, ad-hoc abbreviations and improper casing. On the other hand, clean sources are typically prepared and conformed to certain standards such as those in the academic, journalism and scientific publications. Some common clean sources include news articles from online media sites, and document collections in the form of corpora. Different quality of texts will require different treatments during the preprocessing phase and dirty texts can be much more demanding.

An increasing number of approaches are gearing towards the use of online sources such as corporate intranet [Kietz *et al.*, 2000] and search engines retrieved documents [Cimiano and Staab, 2005] for different aspects of ontology engineering. Despite such growth, recent reviews [Wong *et al.*, 2006] show that only a small number of researchers [Maedche and Volz, 2001; Novacek and Smrz, 2005] acknowledge the effect of text cleanliness on the quality of their ontology engineering output. With the prevalence of online sources, this *"...annoying phase of text cleaning..."*[Mikheev, 2002] has become inevitable and ontology engineering systems can no longer ignore the issue of *text cleanliness*. A recent effort by [Tang *et al.*, 2005] shows that the accuracy of term extraction in text mining improved by 38-45% (F1-measure) with the additional cleaning performed on the input texts (i.e. emails).

Integrated approaches for correcting spelling errors, abbreviations and improper casing have become increasingly appealing as boundaries between different errors in online sources becomes blurred. Along the same line of thought, [Clark, 2003] defended that *"...a unified tool is appropriate because of certain specific sorts of errors"*. To illustrate this idea, consider the error word *"cta"*. Do we immediately take it as a spelling error and correct it as *"cat"*, or is it a problem with the letter casing, which makes it a probable acronym? It is obvious that the problems of spelling error, abbreviation and letter casing are inter-related to a certain extent. The challenge of providing a highly accurate integrated approach for automatically cleaning dirty texts in ontology engineering remains to be addressed.

In an effort to provide an integrated approach to solve spelling errors, ad-hoc abbreviations and improper casing simultaneously, we have developed an **I**ntegrated **S**coring for **S**pelling error correction, **A**bbreviation expansion and **C**ase restoration (ISSAC) [Wong *et al.*, 2006]. The basic ISSAC uses six weights from different sources for automatically correcting spelling error, expanding abbreviations and restoring improper casing. These includes the original rank by the spell checker Aspell [Atkinson, 2006], reuse factor, abbreviation factor, normalized edit distance, domain significance and general significance. Despite the achievements of 96.5% in accuracy by the basic ISSAC, several drawbacks have been identified that call for improvements. In this paper, we present the enhancement of the basic ISSAC. New evaluations performed on seven different set of chat records yield an improved accuracy of 98% as compared to 96.5% and 71% based only on the use of basic ISSAC and of Aspell respectively.

In Section 2, we present a summary of work related to spelling error detection and correction, abbreviation expansion, and other cleaning tasks in general, and also within the context of ontology engineering and text mining. In Section 3, we summarize the basic ISSAC. In Section 4, we propose the enhancement strategies for ISSAC. The evaluation results and discussions are presented in Section 5. We summarize and conclude this paper with future outlook in Section 6.

## 2 Related Work

*Spelling error detection and correction* is the task of recognizing misspellings in texts and providing suggestions for correcting the errors. For example, detecting *"cta"* as an error and suggesting that the error to be replaced with *"cat"*, *"act"* or *"tac"*. More information is usually required to select a correct replacement from a list of suggestions. Two of the most studied classes of techniques are *minimum edit distance* and *similarity key*. The idea of minimum edit distance techniques began with [Damerau, 1964] and [Levenshtein, 1966]. Damerau-Levenshtein distance is the minimal number of insertions, deletions, substitutions and transpositions needed to transform one string into the other. For example, to change *"wear"* to *"beard"* will require a minimum of two operations, namely, a substitution of *'w'* with *'b'*, and an insertion of *'d'*. Many variants were developed subsequently such as the algorithm by [Wagner and Fischer, 1974]. The second class of techniques is the similarity key. The main idea behind similarity key techniques is to map every string into a key such that similarly spelt strings will have identical keys [Kukich, 1992]. Hence, the key, computed for each spelling error, will act as a pointer to all similarly spelt words (i.e. suggestions) in the dictionary. One of the earliest implementation is the SOUNDEX system by [Odell and Russell, 1922]. SOUNDEX is a phonetic algorithm for indexing words based on their pronunciation in English. SOUNDEX works by mapping a word into a key consisting of its first letter followed by a sequence of numbers. For example, SOUNDEX replaces the letter $l_i \in \{A, E, I, O, U, H, W, Y\}$ with $0$ and $l_i \in \{R\}$ with $6$, and hence, $wear \rightarrow w006 \rightarrow w6$ and $ware \rightarrow w060 \rightarrow w6$. Since SOUNDEX, many improved variants were developed such as the Metaphone and the Double-metaphone algorithm by [Philips, 1990], Daitch-Mokotoff Soundex [Lait and Randell, 1993] for Eastern European languages, and others [Holmes and McCabe, 2002]. One of the famous implementation that utilizes the similarity key technique is Aspell [Atkinson, 2006]. Aspell is based on the Metaphone algorithm and the near-miss strategy by its predecessor Ispell [Kuenning, 2006]. Aspell begins by converting the misspelt word to its soundslike equivalent (i.e. metaphone) and moves on to find all words that have a soundslike within one or two edit distances from the original word's soundslike. These soundslike words are the basis for the suggestions of Aspell.

Most of the work in detecting and correcting spelling errors, and expanding abbreviations are carried out separately. The task of *abbreviation expansion* deals with recognizing shorter forms of words (e.g. *"abbr."* or *"abbrev."*), acronyms (e.g. *"NATO"*) and initialisms (e.g. *"HTML"*, *"FBI"*), and expanding them to their corresponding words[1]. The work on detecting and expanding abbreviations are mostly conducted in the realm of named-entity recognition and word-sense disambiguation. The approach presented by [Schwartz and Hearst, 2003] begins with the extraction of all abbreviations and definition candidates based on the adjacency to parentheses. A candidate is considered as the correct definition for an abbreviation if they appears in the same sentence, and the candidate has no more than $min(|A| + 5, |A| * 2)$ words, where $|A|$ is the number of characters in an abbreviation $A$. [Park and Byrd, 2001] presented an algorithm based on rules and heuristics for extracting definitions for abbreviations from texts. Several factors are considered for this purpose such as syntactic cues, priority of rules, distance between abbreviation and definition and word casing. [Pakhomov, 2001] proposes a semi-supervised approach that employs a hand-crafted table of abbreviations and their definitions for training a maximum entropy classifier.

For *case restoration*, improper casing in words are detected and restored. For example, detecting the letter *'j'* in *"jones"* as improper and correcting the word to produce *"Jones"*. [Lita *et al.*, 2003] presented an approach for restoring cases based on the context in which the word exists. The approach first captures the context surrounding a word and approximates the meaning using N-grams. The casing of the letters in a word will depend on the most likely meaning of the sentence. [Mikheev, 2002] presented an approach that identifies sentence boundaries, disambiguate capitalized words and identifying abbreviations using a list of common words and a list of the most frequent words which appear in sentence-starting positions. The approach can be described in four steps: identify abbreviations in texts, disambiguate ambiguously capitalized words, assign unambiguous sentence boundaries and disambiguate sentence boundaries if an abbreviation is followed by a proper name.

In the context of ontology engineering and other related areas such as text mining, spelling errors correction and abbreviations expansion are mainly carried out

---
[1] Some researchers refer to this relationship as *abbreviation and definition* or *short-form and long-form*

as part of the text preprocessing (i.e. text cleaning, text filtering, text normalization) phase. Some other common tasks in text preprocessing include plain text extraction (i.e. format conversion, HTML/XML tag stripping, table identification [Ng et al., 1999]), sentence boundary detection [Stevenson and Gaizauskas, 2000], case restoration [Mikheev, 2002], part-of-speech tagging [Brill, 1992] and sentence parsing [Lin, 1994]. A review by [Wong et al., 2006] shows that nearly all ontology engineering systems in the survey perform only shallow linguistic analysis such as part-of-speech tagging during the text preprocessing phase. These existing approaches require the input to be clean and hence, the techniques for correcting spelling errors, expanding abbreviations and restoring cases are considered as unnecessary. Ontology engineering approaches such as [Xu et al., 2002], Text-to-Onto [Maedche and Volz, 2001] and BOLE [Novacek and Smrz, 2005] are the few exceptions. In addition to shallow linguistic analysis, these systems incorporate some of the cleaning tasks. [Xu et al., 2002] identifies abbreviated variants of proper names (e.g. *HP* for *Hewlett-Packard*) as part of the named-entity tagging process through the use of lexicon. Text-to-Onto extracts plain text from various formats such as PDF, HTML, XML, and identifies and replaces abbreviations using substitution rules based on regular expressions. The text preprocessing phase of BOLE consists of sentence boundary detection, irrelevant sentence elimination and text tokenization using Natural Language Toolkit (NLTK)[2].

In a text mining approach for extracting topics from chat records, [Castellanos, 2003] presented a very comprehensive list of techniques for text preprocessing. The approach employs a thesaurus, constructed using the Smith-Waterman algorithm [Smith and Waterman, 1981], for correcting spelling errors and identifying abbreviations. In addition, the approach removes programming codes from texts based on the characteristics that distinguish codes from normal texts (e.g. shorter lines in program codes, presence of special characters) and detects sentence boundary based on simple heuristics (e.g. punctuation marks followed by an upper case letter). [Tang et al., 2005] presented a cascaded approach for cleaning emails prior to any text mining processing. The approach is composed of four passes: non-text filtering for eliminating irrelevant non-text data such as email header and program code filtering, and sentence normalization, case restoration and spelling error correction for transforming relevant text data into canonical form.

Many of the techniques mentioned above are dedicated to perform only one out of the three different cleaning tasks. In addition, the evaluations conducted to obtain the accuracy are performed in different settings (e.g. no benchmark, test data and agreed measure of accuracy). Hence, it is not possible to compare these different techniques based on the accuracy reported in the respective papers. As pointed out earlier, only a small number of integrated techniques are available for handling all three tasks. Such techniques are usually embedded as part of a larger text preprocessing module. Consequently, the evaluations of the individual cleaning task in such environments are not available.

## 3 Basic ISSAC as Part of Text Preprocessing

ISSAC was initially designed and implemented as part of the text preprocessing phase in an ontology engineering system that uses chat records as input. The use of chat records has required us to place more effort during the text preprocessing phase. Figure 1 highlights the various spelling errors, ad-hoc abbreviations and improper casing that occur much more frequently in chat records than in clean texts.

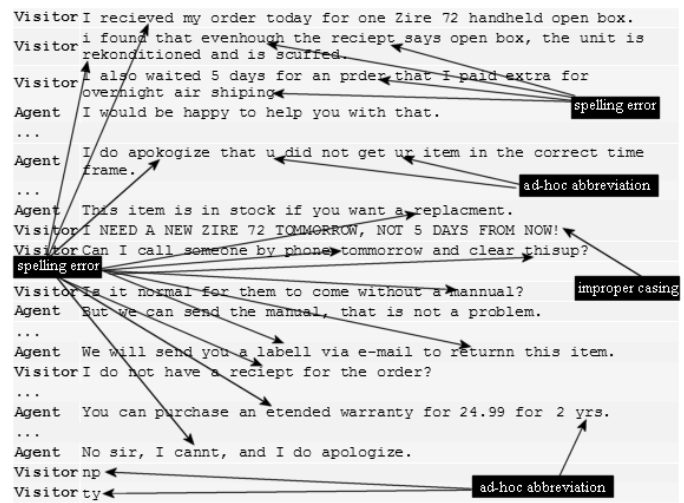

Figure 1: Example of spelling errors, ad-hoc abbreviations and improper casing in a chat record

Prior to spelling error correction, abbreviation expansion and case restoration, three tasks are performed as part of the text preprocessing phase. Firstly, plain text extraction is conducted to remove HTML and XML tags from the chat records using regular expressions and Perl modules, namely, XML::Twig[3] and HTML::Strip[4]. Secondly, identification of URLs, emails, emoticons[5] and tables is performed. Such information is extracted and set aside for assisting in other business intelligence analysis. Tables are removed using signature of a table such as multiple spaces between words and words aligned in columns for multiple lines [Castellanos, 2003]. Thirdly,

---

[2]http://nltk.sourceforge.net/tech/index.html
[3]http://search.cpan.org/~mirod/XML-Twig-3.26/
[4]http://search.cpan.org/~kilinrax/HTML-Strip-1.06/
[5]An emoticon, also called a smiley, is a sequence of ordinary printable characters or a small image, intended to represent a human facial expression and convey an emotion

sentence boundary detection is performed using Lingua::EN::Sentence[6].

Prior to applying ISSAC, each sentence in the input text (e.g. chat record) is tokenized to obtain a list of words $T = \{t_1, ...t_w\}$ which will be fed into Aspell. For each word $e$ that Aspell considers as erroneous, a list of ranked suggestions $S$ is produced. Initially, $S = \{s_{1,1}, ..., s_{n,n}\}$ is an ordered list of $n$ suggestions where $s_{j,i}$ is the $j^{th}$ suggestion with rank $i$ (smaller $i$ indicates higher confidence in the suggested word). If $e$ appears in the *abbreviation dictionary*, the list $S$ is augmented by adding all the corresponding $m$ expansions in front of $S$ as additional suggestions with rank 1. In addition, the error word $e$ is appended at the end of $S$ with rank $n + 1$. These augmentations result in an extended list $S = \{s_{1,1}, ..., s_{m,1}, s_{m+1,1}, ..., s_{m+n,n}, s_{m+n+1,n+1}\}$, which is a combination of $m$ suggestions from the *abbreviation dictionary* (if $e$ is a potential abbreviation), $n$ suggestions by Aspell, and the error word $e$ itself. Placing the error word $e$ back into the list of possible replacements serves one purpose: to ensure that if no better replacement is available, we keep the error word $e$ as it is. Once the extended list $S$ is obtained, each suggestion $s_{j,i}$ is re-ranked using ISSAC. The new score for the $j^{th}$ suggestion with original rank $i$ is defined as

$$NS(s_{j,i}) = i^{-1} + NED(e, s_{j,i}) + RF(e, s_{j,i}) \\ + AF(s_{j,i}) + DS(l, s_{j,i}, r) + GS(l, s_{j,i}, r)$$

where

- $NED(e, s_{j,i}) \in (0, 1]$ is the *normalized edit distance* defined as $(ED(e, s_{j,i}) + 1)^{-1}$ where $ED$ is the minimum edit distance between $e$ and $s_{j,i}$.

- $RF(e, s_{j,i}) \in \{0, 1\}$ is the boolean *reuse factor* for providing more weight to suggestion $s_{j,i}$ that has been previously used for correcting error $e$. The reuse factor is obtained through a lookup into a *history list* that ISSAC keeps to record previous corrections. $RF(e, s_{j,i})$ will provide factor 1 if the error $e$ has been previously corrected with $s_{j,i}$ and 0 otherwise.

- $AF(s_{j,i}) \in \{0, 1\}$ is the *abbreviation factor* for denoting that $s_{j,i}$ is a potential abbreviation. A lookup into the *abbreviation dictionary*, $AF(s_{j,i})$ will yield factor 1 if suggestion $s_{j,i}$ exists in the dictionary and 0 otherwise. When the scoring process takes place and the corresponding expansions for potential abbreviations are required, www.stands4.com is consulted. A copy of the expansion is stored in a local *abbreviation dictionary* for future reference.

- $DS(l, s_{j,i}, r) \in [0, 1]$ measures the *domain significance* of suggestion $s_{j,i}$ based on its appearance in the domain corpora by taking into account the neighbouring words $l$ and $r$. The weight is defined as the ratio between the frequency of occurrence of $s_{j,i}$ (individually, and within $l$ and $r$) in the domain corpora and the sum of the frequencies of occurrences of all suggestions (individually, and within $l$ and $r$).

- $GS(l, s_{j,i}, r) \in [0, 1]$ measures the *general significance* of suggestion $s_{j,i}$ based on its appearance in the general collection (e.g. Goggle retrieved documents). The idea behind general significance is similar to that of domain significance. The weight is defined as the ratio between the number of documents in the general collection containing $s_{j,i}$ within $l$ and $r$ and the number of documents in the general collection that contains $s_{j,i}$ alone. Both the ratios in $DS$ and $GS$ are offset by a measure similar to that of the *Inverse Document Frequency*. For further details of $DS$ and $GS$, please refer to [Wong et al., 2006].

## 4 Enhancement of ISSAC

The list of suggestions and the initial ranks provided by Aspell are integral parts of ISSAC. Table 1 summarizes the accuracy of basic ISSAC obtained from the previous evaluations [Wong et al., 2006] on four sets of chat records. The achievement of 74.4% accuracy by Aspell from the previous evaluations, given the extremely poor nature of the texts, demonstrates the strength of the Metaphone algorithm and near-miss strategy. The further increase of 22% in accuracy using basic ISSAC demonstrates the potential of the combined weights $NS(s_{j,i})$.

**Table 1. Accuracy of basic ISSAC from previous evaluations**

|  | Evaluation 1 | Evaluation 2 | Evaluation 3 | Evaluation 4 | Average |
|---|---|---|---|---|---|
| number of correct replacements using ISSAC | 97.06% | 97.07% | 95.92% | 96.20% | 96.56% |
| number of correct replacements using Aspell | 74.61% | 75.94% | 71.81% | 75.19% | 74.39% |

Based on the results of the previous evaluations, we have discussed in detail the three causes behind the remaining 3.5% of errors which have been wrongfully replaced. Table 2 shows the breakdown of the causes behind the errors with wrong replacement by the basic ISSAC.

**Table 2. The breakdown of the causes behind the errors with wrong replacement by basic ISSAC**

| Causes | Basic ISSAC |
|---|---|
| Correct replacement not in suggestion list | 2.00% |
| Inadequate/erroneous neighbouring words | 1.00% |
| Anomalies | 0.50% |

The three causes are summarized as follow:

1. The accuracy of correction by basic ISSAC is bounded by the coverage of the list of suggestions

---
[6] http://search.cpan.org/dist/Lingua-EN-Sentence/

$S$ produced by Aspell. About 2% of wrong replacements is due to the absence of the correct replacement from the list of suggestions produced by Aspell. For example, the error *"prder"* in the context of *"The prder number"* was wrongly replaced by both Aspell and basic ISSAC as *"parader"* and *"prder"* respectively. After a look into the evaluation log, we realized that the correct replacement *"order"* was not in $S$.

2. The use of the two immediate neighbouring words $l$ and $r$ to inject more contextual consideration into domain and general significance has contributed to a large portion of the increase in accuracy. Nonetheless, the use of $l$ and $r$ in ISSAC is by no means perfect. About 1% out of the total errors with wrong replacement is due to two flaws related to $l$ and $r$, namely, neighbouring words with incorrect spelling, and neighbouring words who are *inadequate*. When the neighbouring words are incorrect, $DS$ and $GS$ will fail to capture the actual significance of the correct replacement with respect to the erroneous left or right word. The neighbouring words are considered as inadequate due to their *indiscriminative* nature. For example, the left word *"both"* in *"both ocats are"* does not provide much clue as to adequately discriminate between suggestions such as *"coats"*, *"cats"* and *"acts"*.

3. The remaining 0.5% can be seen as anomalies where basic ISSAC does not address. There are two cases of anomalies: the equally likely nature of all the possible replacements, and the *contrasting* value of certain weights. As an example for the first case, consider the error *"Janice cheung <"*. The left word is correctly spelt and has adequately confined the suggestions to proper names. In addition, the correct replacement *"Cheung"* is present as a suggestion $s_{j,i} \in S$. Despite all these, both Aspell and ISSAC decided to replace *"cheung"* with *"Cheng"*. A look into the evaluation log reveals that the surname *"Cheung"* is as common as *"Cheng"*. In such cases, the probability of replacing $e$ with the correct replacement is $c^{-1}$ where $c$ is the number of suggestions with approximately same $NS(s_{j,i})$. The second case of anomalies is due to contrasting value of certain weights, especially $NED$ and $i^{-1}$, that causes wrong replacements. For example, in the case *"cannot chage an"*, basic ISSAC replaced the error *"chage"* with *"charge"* instead of *"change"*. All the other weights for *"change"* are comparatively higher (i.e. $DS$ and $GS$) or the same (i.e. $RF$, $NED$ and $AF$) as *"charge"*. Such inclination indicates that *"change"* is the most proper replacement given the various cues. Nonetheless, the original rank by Aspell for *charge* is *i=1* while *change* is *i=6*. As smaller $i$ indicates higher confidence, the inverse of the original rank by Aspell $i^{-1}$ results in the plummeting of the combined weight for *"change"*.

In this paper, we attempt to approach the enhancement of ISSAC from the perspective of the first and second cause. For this purpose, we proposed three modifications to the basic ISSAC:

1. We proposed the use of additional spell checking facilities as the answer to the first cause (i.e. compensating the inadequacy of Aspell). Google spellcheck, which is based on statistical analysis of words on the World Wide Web[7], appears to be the ideal candidate for complementing Aspell. Using the Google SOAP API[8], we can have easy access to one of the many functions provided by Google, namely, Google *spelling requests*. Our new evaluations show that Google spellcheck works well for many errors when Aspell fails to suggest the correct replacement. Similar to adding the expansions for abbreviations and the suggestions by Aspell, the suggestion for an error provided by Google is added in front of the list of all suggestions $S$ with rank 1. This will place the suggestion by Google on the same rank as the first suggestion by Aspell, and let ISSAC determines the most suitable replacement.

2. The basic ISSAC relies on only Aspell for determining if a word is an error. For this purpose, we decided to include Google spellcheck as a complement. If a word is detected as possible error by either Aspell or Google spellcheck, then we have adequate evidence as to proceed to the process of correcting it using enhanced ISSAC. In addition, errors that result in valid words are not recognized by Aspell. For example, Aspell will not recognize *"hat"* as an error. If we were to take into consideration the neighbours that it co-occurs with, namely, *"suret hat they"*, then *"hat"* is certainly an error. Google contributes in this aspect. In addition, the use of Google spellcheck has also indirectly provided ISSAC with a partial solution to the second cause (i.e. erroneous neighbouring words). Whenever Google is checking a word for spelling error, the neighbouring words are simultaneously examined. For example, while providing suggestion to the error *"tha"*, Google will simultaneously take into consideration the neighbours, namely, *"sure tha tthey"*, and suggest that the right word *"tthey"* be replaced with *"they"*. Google's ability to consider contextual information is empowered by the statistical evidence gathered in the form of co-occurrences of two terms in documents on the World Wide Web. Pairs of terms are ruled out as statistically improbable when their co-occurrences are extremely low. In such cases, Google can confidently attempt to suggest a better partner for the term.

3. We have altered the *reuse factor RF* by eliminating the use of *history list* that gives more weight to suggestions that have been previously chosen to correct particular errors. We have come to realize that there is no guarantee a particular replacement for an error is correct. When a replacement is incorrect and is stored in the *history list*, the *reuse factor* will propagate the wrong replacement to the subsequent corrections. Therefore, we adapted the *reuse factor* to support the use of Google spellcheck in the form of entries in a local *spelling dictionary*. There are two types of entries in the *spelling dictionary*. The main type is the suggestions by Google for spelling errors. This type of entries is automatically updated every time Google suggest a replacement for an error. The second type, which is optional, is the suggestions for errors that are manually entered by users. Hence the modified *reuse factor* will now give the weight of 1 to only suggestions that are provided by Google spellcheck or predefined by users.

---

[7]http://www.google.com/help/features.html
[8]http://www.google.com/apis

Despite the certain level of superiority that Google spellcheck exhibits in the three enhancements, Aspell remains necessary. Google spellcheck is based on the occurrences of words on the World Wide Web. Determining whether a word is an error or not depends very much on its popularity. Even if a word does not exist in the English dictionary, Google will not judge it as an error as long as its popularity exceeds some threshold set by Google. This popularity approach has both its pros and cons. On the one hand, such approach is good for recognizing proper nouns, especially emerging ones, such as "iPod" and "Xbox". On the other hand, words such as "thanx" in the context of "[ok] [thanx] [for]" is not considered as an error by Google even though it should be corrected.

The algorithm for text preprocessing that comprises of the basic ISSAC together with all its enhancements is described in Algorithm 1.

## 5 Evaluation and Discussion

Evaluations are conducted using chat records provided by `247Customer.com`[9]. As a provider of customer lifecycle management services, the chat records by `247Customer.com` offer a rich source of domain information in a natural setting (i.e. conversations between customers and agents). Consequently, these chat records are filled with spelling errors, ad-hoc abbreviations, improper casing and many other problems that are considered as intolerable by many of the existing language and speech applications. Therefore, these chat records become the ideal source for evaluating ISSAC. Four sets of test data, each comes in an XML file of 100 chat sessions, were employed in the previous evaluations [Wong *et al.*, 2006]. To evaluate the enhanced ISSAC, we have included an additional three sets which brings the total number of chat records to 700. The chat records and Google constitutes the domain corpora and general collection respectively during the evaluation. GNU Aspell version 0.60.4 [Atkinson, 2006] is employed for detecting errors and generating suggestions.

---

[9]http://www.247customer.com/

---

**Algorithm 1** Enhanced ISSAC

1: **input**: chat records or other online documents
2: Remove all HTML or XML tags from input documents
3: Extract and keep URLs, emails, emoticons and tables
4: Detect and identify sentence boundary
5: **for** each document **do**
6:   **for** each sentence in the document **do**
7:     tokenize the sentence to produce a set of words $T = \{t_1, ..., t_w\}$
8:     **for** each word $t \in T$ **do**
9:       Identify left $l$ and right $r$ word for $t$
10:       **if** $t$ consists of all upper case **then**
11:         Turn all letters in $t$ to lower case
12:       **else if** $t$ consists of all digits **then**
13:         **next**
14:       **end if**
15:       Feed $t$ to Aspell
16:       **if** $t$ is identified as error by Aspell or Google spellcheck **then**
17:         **initialize** $S$ and $NS$, the set of suggestions for error $t$, and an array of new scores for all suggestions for error $t$ respectively
18:         Add the $n$ suggestions for word $t$ produced by Aspell to $S$ according to the original rank from 1 to $n$
19:         Perform a lookup in the *abbreviation dictionary* and add all the corresponding $m$ expansions for $t$ at the front of $S$, all with rank 1
20:         Perform a lookup in the *spelling dictionary* and add the retrieved suggestion at the front of $S$ with rank 1
21:         Add the error word $t$ itself at the end of $S$, with rank $n+1$
22:         The final $S$ is $\{s_{1,1}, s_{2,1}, ..., s_{m+1,1}, s_{m+2,1}, ..., s_{m+n+1,n}, s_{m+n+2,n+1}\}$ where $j$ and $i$ in $s_{j,i}$ is the element index and the rank respectively
23:         **for** each suggestion $s_{j,i} \in S$ **do**
24:           Determine $i^{-1}$, $NED$ between error $e$ and the $j^{th}$ suggestion, $RF$ by looking into the *spelling dictionary*, $AF$ by looking into the *abbreviation dictionary*, $DS$, and $GS$
25:           Sum the weights and push the sum into $NS$
26:         **end for**
27:         Correct word $t$ with the suggestion that has the highest combined weights in array $NS$
28:       **end if**
29:     **end for**
30:   **end for**
31: **end for**
32: **output**: documents with spelling errors corrected, abbreviations expanded and improper casing restored.

Similar to the previous evaluations, determining whether a suggestion by either Aspell or enhanced ISSAC as a correct replacement for an error is a delicate process that must be performed manually. For example, it is difficult to automatically determine whether error *"itme"* should be replaced with *"time"* or *"item"* without more information (e.g. the neighbouring words). The evaluation of the errors and replacements are conducted in a unified manner. The errors are not classified into spelling errors, ad-hoc abbreviations or improper casing. For example, should the error *"az"* (*"AZ"* is the abbreviation for the state of *"Arizona"*) in the context of *"Glendale az <"* be considered as an abbreviation or improper casing? The boundaries between the different types of *dirtiness* that occur in real-world texts, especially those from online sources, are not clear. This is the main reason behind the increasing number of efforts that attempt to provide techniques to handle various *dirtiness* in an integrated manner [Sproat *et al.*, 2001; Mikheev, 2002; Clark, 2003; Tang *et al.*, 2005].

**Table 3a. Accuracy of enhanced ISSAC over seven evaluations**

|  | Evaluation 1 | Evaluation 2 | Evaluation 3 | Evaluation 4 |
|---|---|---|---|---|
| number of correct replacements using enhanced ISSAC | *98.45%* | *97.91%* | *98.40%* | *98.23%* |
| number of correct replacements using basic ISSAC | *97.06%* | *97.07%* | *95.92%* | *96.20%* |
| number of correct replacements using Aspell | *74.61%* | *75.94%* | *71.81%* | *75.19%* |

**Table 3b. Accuracy of enhanced ISSAC over seven evaluations**

|  | Evaluation 5 | Evaluation 6 | Evaluation 7 | Average |
|---|---|---|---|---|
| number of correct replacements using enhanced ISSAC | *97.39%* | *97.85%* | *97.86%* | *98.01%* |
| number of correct replacements using basic ISSAC | *95.64%* | *96.65%* | *97.14%* | *96.53%* |
| number of correct replacements using Aspell | *63.62%* | *65.79%* | *70.24%* | *71.03%* |

After a careful evaluation of all replacements suggested by Aspell and by enhanced ISSAC for all 3313 errors, we discovered a further improvement in accuracy using the latter. As shown in Table 3a and 3b, the use of the first suggestion by Aspell as replacement for spelling errors yields an average of 71%, which is a decrease from 74.4% in the previous evaluations due to the additional *dirtiness* in the extra three sets of chat records. With the addition of the various weights that form basic ISSAC, an average increase of 22% was achieved, resulting to an improved accuracy of 96.5%. As predicted, the enhanced ISSAC score a much better accuracy at 98%.

The increase of 1.5% from basic ISSAC is contributed by the suggestions from Google that complement the inadequacies of Aspell. A previous error *"prder"* within the context of *"The prder number"* that could not be corrected by basic ISSAC due to the first cause was solved after our enhancements. The correct replacement *"order"* was suggested by Google. Another error *"ffer"* in the context of *"youo ffer on"* that could not be corrected due to the second cause was successfully replaced by *"offer"* after Google has simultaneously corrected the left word to *"you"*.

The increase of accuracy by 1.5% is in line with the drop in the number of errors with wrong replacements due to the absence of correct replacements from suggestions by Aspell, and the erroneous neighbouring words. As shown in Table 4, there is a visible drop in the number of errors with wrong replacements due to the first and second cause from the existing 2% (as shown in Table 1) to 0.8%, and 1% (as shown in Table 1) to 0.7% respectively.

**Table 4. The breakdown of the causes behind the errors with wrong replacement by enhanced ISSAC**

| Causes | Enhanced ISSAC |
|---|---|
| Correct replacement not in suggestion list | 0.80% |
| Inadequate/erroneous neighbouring words | 0.70% |
| Anomalies | 0.50% |

# 6 Conclusion

As an increasing number of ontology engineering systems are opening up to the use of online sources, the need to handle *dirty texts* becomes inevitable. Regardless of whether we acknowledge this fact, the quality of ontology and the proper functioning of the systems are, to a certain extent, dependent on the cleanliness of the input texts. Most of the existing techniques for correcting spelling errors, expanding abbreviations and restoring cases are studied separately. We, along with an increasing number of researchers, have acknowledged the fact that many *dirtiness* in texts are composite in nature (i.e. multi-error). As we have demonstrated during our evaluation and discussion in this paper, many errors are difficult to be classified as either spelling errors, ad-hoc abbreviations or improper casing.

In this paper, we present the enhancement of the **I**ntegrated **S**coring for **S**pelling error correction, **A**bbreviation expansion and **C**ase restoration (ISSAC). The basic ISSAC was build upon the famous spell checker Aspell for simultaneously providing solution to spelling errors, abbreviations and improper casing. This scoring mechanism combines weights based on various information sources, namely, original rank by Aspell, reuse factor, abbreviation factor, normalized edit distance, domain significance and general significance. In the course of evaluating basic ISSAC, we have discovered and discussed in detail three causes behind the errors with wrong correction. We approached the enhancement of ISSAC from the first and the second cause, namely, the absence of the correct replacement from the suggestions by Aspell, and the inadequacies of the neighbouring words. We proposed three modifications to the basic ISSAC, namely, 1) the use of Google spellcheck for compensating the inadequacy of Aspell, 2) the incorporation of Google spellcheck for determining if a word is erroneous, and 3) the alteration of the *reuse factor RS* by shifting from the use of a *history list* to a *spelling dictionary*. Evaluations performed using the enhanced ISSAC on seven sets of chat records revealed a further improved accuracy at 98% from the previous 96.5% using basic ISSAC.

Even though the idea for ISSAC was first motivated and conceived within the paradigm of ontology engineering, we see great potentials in further improvements and fine-tuning for a wide range of uses, especially in language and speech applications. We hope that a unified approach such as ISSAC will pave the way for more research in providing a complete solution for text preprocessing (i.e. text cleaning) in general.

## Acknowledgement

This research was supported by the Australian Endeavour International Postgraduate Research Scholarship, and the Research Grant 2006 by the University of Western Australia. The authors would like to thank 247Customer.com for providing the evaluation data. Gratitude to the developer of GNU Aspell, Kevin Atkinson.